\documentclass[conference] {IEEEtran}
\usepackage{graphicx} 
\usepackage{amsmath}
\usepackage{amsfonts}
\usepackage{amsmath,latexsym,amsfonts,enumerate}
\usepackage{amssymb,amsthm}
\usepackage{mathtools}
\usepackage{csquotes}
\usepackage{tikz-cd}
\usepackage[ruled,vlined]{algorithm2e}
\usepackage{tikz}
\usepackage{comment}
\usepackage[T1]{fontenc}
\usepackage{cleveref}
\usepackage{lmodern}
\usepackage{algpseudocode}
\usepackage{graphicx}
\usepackage{enumitem}

\usepackage[total={6in,10in},left=1.5in,top=0.5in,includehead,includefoot]{geometry}
\usepackage{microtype}
\usepackage[nodisplayskipstretch]{setspace}
\usepackage{amsmath}
\usepackage{amsfonts}
\usepackage{mathtools}
\usepackage{tensor}
\everymath{\displaystyle}

\DeclareMathOperator{\argmax}{argmax}

\newtheorem{theor}{Theorem}

\newtheorem*{ques*}{Question}
\newtheorem*{theor*}{Theorem}
\newtheorem*{lem*}{Lemma}
\newtheorem*{prop*}{Proposition}
\newtheorem*{cor*}{Corollary}
\newtheorem*{defin*}{Definition}
\newtheorem*{rem*}{Remark}
\newtheorem*{rems*}{Remarks}
\newtheorem*{ex*}{Example}
\newtheorem*{notat*}{Notations}
\newtheorem*{claim*}{Claim}

\title{Active Learning Classification from a Signal Separation Perspective}
\author{}
\date{February 10, 2025}

\newcommand{\ignore}[1]{}

\begin{document}

\author{
\IEEEauthorblockN{Hrushikesh Mhaskar\IEEEauthorrefmark{1}\IEEEauthorrefmark{2}, Ryan O'Dowd\IEEEauthorrefmark{2} and Efstratios Tsoukanis\IEEEauthorrefmark{2}}
\IEEEauthorblockA{ \\ 
Institute of Mathematical Sciences, Claremont Graduate University,  Claremont, CA, 91711\\
Emails: Hrushikesh.Mhaskar@cgu.edu, ryan.o'dowd@cgu.edu, efstratios.tsoukanis@cgu.edu
}
}

\maketitle

\begingroup\renewcommand\thefootnote{\IEEEauthorrefmark{1}}
\footnotetext{Supported in part by ONR grant N00014-23-1-2394.}
\endgroup

\begingroup\renewcommand\thefootnote{\IEEEauthorrefmark{2}}
\footnotetext{Supported in party by ONR grant N00014-23-1-2790.}
\endgroup

\begin{abstract}
In machine learning, classification is usually seen as a function approximation problem, where the goal is to learn a function that maps input features to class labels.
In this paper, we propose a novel clustering and classification framework inspired by the principles of signal separation. This approach enables efficient identification of class supports, even in the presence of overlapping distributions. We validate our method on real-world hyperspectral datasets Salinas and Indian Pines. The experimental results demonstrate that our method is competitive with the state of the art active learning algorithms by using a very small subset of data set as training points.
\end{abstract}

\section{Introduction}\label{sec:into}

Classification is one of the oldest and most extensively studied problems in machine learning. Mathematically, the problem can be formulated as follows: we consider data of the form \( \{(x, y)\} \), where \( x \in \mathcal{D} \) for some domain \( \mathcal{D} \) (e.g., Euclidean space or graph vertices), and \( y \) takes values in a finite set, conveniently encoded as \( \{1, \dots, K\} \) for some integer \( K \ge 2 \). In the pair \( (x, y) \), \( y \) is the \textbf{class label} of \( x \), and the \textbf{classification function} is defined as \( f(x) = y \).

Over the past 50 years, machine learning research has produced numerous algorithms for estimating the class label of any data point \( x \in \mathcal{D} \), all of which approximate the function \( f \). This perspective unifies the classification problem with the classical problem of function approximation.

\textbf{Approximation theory} focuses on methods for estimating real-valued functions defined on subsets of Euclidean space and analyzing the intrinsic (non-statistical) errors associated with such approximations. Typically, the target function \( f \) is assumed to be “smooth” on its domain. Although the classification function is inherently piecewise constant and discontinuous, this is not a theoretical obstacle when classes are well-separated. In such cases, extension theorems by Stein \cite{stein1970singular} guarantee the existence of infinitely differentiable extensions of piecewise constant functions to the entire Euclidean space, even preserving the magnitude of the derivatives.

In modern applications, classes often overlap, and even when they are disjoint, their boundaries may lack smoothness, posing challenges for classical function approximation techniques in classification. Additionally, extension theorems provide only existence results without offering constructive methods to obtain such extensions, particularly when class boundaries are unknown. We propose that classification can be effectively approached through an analogy to the problem of signal separation in phased array antennas.

In this problem, one wants to determine a linear combination $\mu=\sum_{k=1}^K a_k\delta_{\omega_k}$, (where $\delta_\omega$ denotes the Dirac delta distribution supported at $\omega$), given the Fourier coefficients $\hat{\mu}(\ell)=\sum_{k=1}^K a_k\exp(-i\ell \omega_k)$ for finitely many values of $\ell$,
say $|\ell|<n$ for some integer $n$. One way to solve this problem is to consider a smooth, even, low pass filter $H$ supported on $[-1,1]$, and construct
$$
\sigma_n(\mu)(x)=\sum_{\ell\in\mathbb{Z}}H(|\ell|/n)\hat{\mu}(\ell)\exp(i\ell x).
$$

Under certain conditions, we have shown in \cite{mhaskar2024robust,singdet,trigwave,loctrigwave} that $\sigma_n(\mu)\approx \mu$, so that the set $\{ x: |\sigma_n(\mu)(x)|>\min_k|a_k|/2\}$ splits into exactly $K$ clusters and the maxima of $|\sigma(\mu)(x)|$ in these clusters occur very close to the points $\omega_k$.
This is because $\sigma_n(\mu)$ is a convolution of $\mu$ with the kernel $\Phi_n^T(y)=\sum_\ell H(|\ell|/n)\exp(iky)$, and this kernel is highly localized when $H$ is smooth. 

To connect with classification, assume \( a_k \ge 0 \) for all \( k \) with \( \sum_k a_k = 1 \). Suppose the data belongs to classes \( \{\omega_1, \dots, \omega_K\} \), where \( x = \omega_k \) implies class label \( k \). Here, \( \mu \) represents the data distribution, and each class \( k \) is sampled from \( \delta_{\omega_k} \).

Generalizing, each class \( k \) corresponds to a distribution \( \mu_k \), making the overall data distribution a convex combination:
\[
\mu^* = \sum_{k=1}^K a_k \mu_k.
\]
The technical barriers in this viewpoint are the following: (1) the supports of \( \mu_k \) may be continuous rather than discrete, and (2) we observe random samples from \( \mu^* \) instead of Fourier coefficients.

We propose a framework using localized kernels based on Chebyshev polynomials to address classification analogously to signal separation. Our approach builds on \cite{cloninger2020cautious} offering a solid theoretical basis. Though detailed proofs are omitted, extensive experiments highlight the effectiveness of our algorithm, further enhanced by an iterative active learning process across diverse datasets.

\section{Related Work}

Active learning has been explored extensively to improve learning efficiency. Settles \cite{settles2009active} provides a comprehensive survey of active learning strategies, while Dasgupta \cite{dasgupta2011two} offers theoretical insights into its effectiveness.

In super-resolution, Candès and Fernandez-Granda \cite{candes2014towards} introduced convex optimization techniques for recovering point sources, and Tang et al. \cite{tang2013compressed} applied compressed sensing to spectral estimation.

Cloninger and Mhaskar \cite{cloninger2020cautious} introduced cautious active clustering using localized kernels based on Hermite polynomials.
Our approach requires a smaller number of data points.

\section{Theoretical Framework}

Let $\mu^*$ denote a probability measure supported on $\mathbb{X} \subset \mathbb{R}^q$, representing the distribution of a data  set. 
With an appropriate stereographic projection, we may in fact assume that $\mathbb{X}\subset \mathbb{S}^q$, where $\mathbb{S}^q$ denotes the unit sphere of $\mathbb{R}^{q+1}$.
We denote by $\mathbb{B}(x,r)$ the spherical cap of radius $r$, centered at $x\in \mathbb{S}^q$, and for any subset $A\subseteq \mathbb{S}^q$, write $\mathbb{B}(A,r)=\cup_{x\in A}\mathbb{B}(x,r)$. 
We say that $\mu^*$ is \textbf{detectable} if there exists $\alpha>0$ such that for every $x\in \mathbb{S}^q$,
\begin{equation}
\label{eq:measurecond}
\begin{aligned}
\mu^*(\mathbb{B}(x,r)\le c_1r^\alpha, & \mbox{ for $r>0$},\\
\mu^*(\mathbb{B}(x,r)\ge c_2r^\alpha, & \mbox{ for $0<r\le 1$}.
\end{aligned}
\end{equation}
It is hoped that the support $\mathbb{X}$ should have lower dimension than the ambient dimension $q$. 
The first inequality is satisfied, for example, if $\mathbb{X}$ is a submanifold of dimension $\alpha$ and $\mu^*$ is its Riemannian volume measure.
The second inequality above is analogous to the minimal magnitude of the coefficients in the signal separation problem, and plays the same role in our theory.

We assume that there are finitely many ($K$) classes in the data set, with the $k$-th class arising from a probability distribution $\mu_k$.
Thus, we assume that $\mu^*$ is a convex combination of $\mu_k$'s. 
Ideally, the support of the measures $\mu_k$ should be disjoint.
To allow for an overlap of class boundaries, we assume instead that for any $\eta>0$, the support $\mathbb{X}$ of $\mu^*$ is a disjoint union of $K_\eta$ sets $S_{k,\eta}$, separated by a threshold $\eta>0$ and an extra set, $S_{K_\eta+1}$ representing the overlaps. 
We assume that $\mu^*(S_{K_\eta+1})\to 0$ as $\eta\to 0+$.
We will say that $\mu^*$ has a \textbf{fine structure} if it is detectable and such a partition exists.
Our goal is to separate these sets.

Towards this goal, we will use the localized polynomial defined on $[-1,1]$ by
\begin{equation}
\label{eq:kerndef}
\Phi_n(\cos\theta)=1+2\sum_{\ell=1}^{n-1}H\left(\frac{\ell}{n}\right)\cos(\ell\theta),  \qquad \theta\in [0,\pi].
\end{equation}
where the degree $n$ is a tunable parameter, $H: [0,\infty)\to [0,1]$ is infinitely differentiable and non-increasing function, such that $H(t)=1$ if $t\in [0,1/2]$ and $H(t)=0$ if $t\ge 1$. 
It is easy to prove using the Poisson summation formula \cite{singdet} that for any $S\ge 2$, there exists a constant $c=c(H,S)>0$ such that
\begin{equation}
\label{eq:localization}
|\Phi_n(\cos\theta)| \le \frac{cn}{\max(1, (n\theta)^S)}, \qquad \theta\in [0,\pi], \ n\ge 2.
\end{equation}
We note that if $x, y\in \mathbb{S}^q$, then $\Phi_n(\langle x, y\rangle)$ is a spherical polynomial in both $x$ and $y$. Since the geodesic distance between $x$ and $y$ is given by $\rho(x,y)=\arccos(\langle x, y\rangle)$, Eqn.~\eqref{eq:localization} becomes
\begin{equation}
\label{eq:sphlocalization}
|\Phi_n(\langle x, y\rangle)| \le \frac{cn}{\max(1, (n\rho(x,y))^S)}, \ x,y \in \mathbb{S}^q, \ n\ge 2.
\end{equation}

 \section{Main results}
In this section, we introduce the main theorems of this paper. We start by defining our measure estimator:

\begin{equation}
    F_{n,M}(x) := \frac{1}{M}\sum_{j=1}^{M} \Phi_n(\langle x, x_j\rangle)^2.
\end{equation}

We use this  estimator to generate sets which approximate the support of $\mu^*$. In particular, we define:

\begin{equation}\label{eq:eqGn}
\scriptsize
\begin{aligned}
    \mathcal{G}_n(\Theta) := \Big\{ x \in \mathbb{S}^q 
     F_{n,M}(x) \ge \Theta \max_{1 \leq k \leq M} F_{n,M}(x_k) \Big\}.
\end{aligned}
\end{equation}

The following theorem is proved in \cite{Mhaskar2025Sep} 
\begin{theor}\label{theor1}  
\textit{Let $\mu^*$ be a probability measure with a fine structure given by parameter $\eta$ and $S\ge q+2$ be an integer. Let $M \ge c_3n^\alpha \log(n)$ and $\{x_1, x_2, \ldots, x_M\}$ be independent samples from $\mu^*$. 
 There exists $r(\Theta) \sim \Theta^{-1/(S-\alpha)}$ such that with probability at least $1 - c_4/M^{c_5}$ we have}
\begin{equation}
    \mathbb{X} \subseteq \mathcal{G}_n(\Theta) \subseteq \mathbb{B}(\mathbb{X}, r(\Theta)/n).
\end{equation}
\textit{Moreover, if $n>2r(\Theta)/\eta$ there exists a partition $\{\mathcal{G}_{k,\eta,n}(\Theta)\}_{k=1}^{K_\eta+1}$ of $\mathcal{G}_n(\Theta)$ with the following properties:}
\begin{equation}\label{eq:clusterseparate}
    \text{dist}(\mathcal{G}_{j,\eta,n}(\Theta), \mathcal{G}_{k,\eta,n}(\Theta)) \geq \eta, \quad j \neq k,
\end{equation}
\textit{and}
\begin{equation}\label{eq:componentapprox}
    \mathcal{S}_{k,\eta} \subseteq \mathcal{G}_{k,\eta,n}(\Theta) \subseteq \mathbb{B}(\mathcal{S}_{k,\eta}, r(\Theta)/n).
\end{equation}
\end{theor}
\section*{Brief Overview of Our Algorithm}

The \textbf{Salinas} and \textbf{Indian Pines} hyperspectral datasets are well-known benchmarks in image analysis, commonly used to assess classification algorithms due to their rich spectral data.
The \text{Salinas dataset}, captured by the  AVIRIS over California's Salinas Valley, provides high spatial resolution with \text{3.7-meter pixels} and \text{224 spectral bands} covering the \text{0.4--2.5 µm} range. It mainly features agricultural areas, making it ideal for land cover and vegetation classification. For our experiments, we use the first \text{10 classes} out of 13, selecting $50\%$ of the data from each class randomly.

The \text{Indian Pines dataset}, also from AVIRIS over \text{northwestern Indiana}, consists of a \text{$145 \times 145$ pixel grid} with \text{220 spectral bands}. It poses a classification challenge due to spectral overlap among different land covers. We analyze a \text{$57 \times 41$ pixel subset} containing \text{corn-notill}, \text{stone-steel-towers}, \text{woods}, \text{soybean-mintil}, and \text{grass-trees}, with each pixel having \text{220 spectral features}.

Our algorithm, implements \Cref{theor1} with the following modification. We use an \textbf{iterative refinement process} that optimizes clustering by dynamically adjusting kernel parameters and thresholds, validated on real-world datasets like \textbf{Salinas} and \textbf{Indian Pines}.

We first apply PCA to reduce data from $\mathbb{R}^d$ to $\mathbb{R}^{d'}$ ($d' < d$), preserving maximum variance. The transformed data is normalized and projected onto the unit hypersphere $S^q$.

Next, we compute the \textbf{angle matrix}:
\[
A_{ij} = \arccos(\langle x_i, x_j \rangle),
\]
quantifying angular distances between \(x_i, x_j \in S^q\). 

We build an \textbf{adjacency graph} \(G = (V, E)\) with edges:
\[
E = \{ (x_i, x_j) : A_{ij} < \eta \},
\]
and identify connected components \(\{ C_{n,\ell} \}\). The refinement process adjusts kernel degree \(n\) and threshold \(\Theta\) to stabilize clustering. Class supports are approximated by $G_n(\Theta)$.

For uncertain components, we iteratively query the most confident points, propagate labels within components, and update \(n\) and \(\Theta\) until labeling stabilizes.

In \textbf{post-processing}, we use the \textbf{Witness Function Method} as in \cite{witnesspaper} to propagates labels except instead of  Hermite based polynomial kernels in \cite{witnesspaper} we use the following kernel introduce in \cite{mhaskar2024learning}:
The matrix \( \Phi_{n,q} \) is defined as:

{\scriptsize\[
\Phi_{n,q}(x) = \sum_{k=0}^{n-1} H\left(\frac{k}{n}\right) \frac{P_k^{\left(\frac{q}{2}-1, \frac{q}{2}-1\right)}(1) \, P_k^{\left(\frac{q}{2}-1, \frac{q}{2}-1\right)}(x)}{N_k}
\]}

where \( P_k^{(\alpha, \beta)}(x) \) are Jacobi polynomials with \( \alpha = \beta = \frac{q}{2} - 1 \), and the normalization factor \( N_k \) is:

{\scriptsize
\[
N_k = 2^{\alpha + \beta + 1} \frac{\Gamma(k + \alpha + 1) \Gamma(k + \beta + 1)}{\Gamma(k + 1) \Gamma(k + \alpha + \beta + 1)} \cdot \frac{1}{2k + \alpha + \beta + 1}
\]
}

{\scriptsize
\[
\hat{y}(x) = \arg\max_{k} \sum_{x_i \in \mathcal{A}_k} \Phi_{n,q}(\langle x, x_i \rangle),
\]
}
where \(\mathcal{A}_k\) denotes confidently labeled points of class \(k\).

\begin{figure}[h]
    \centering
    \includegraphics[width=0.4\textwidth]{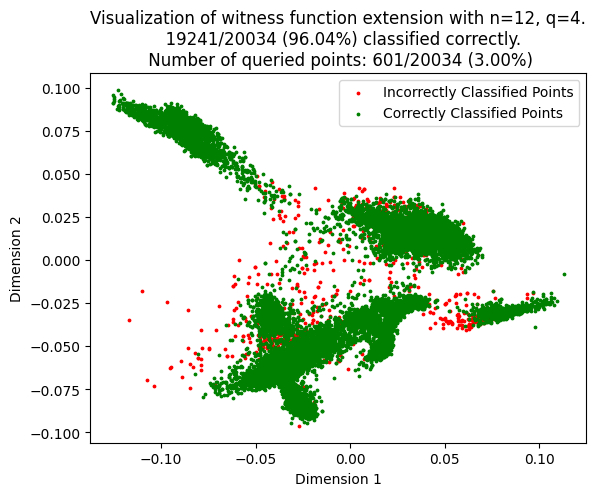}
    \caption{Salinas dataset.}
    \label{Salinas}
\end{figure}

As shown in \Cref{Salinas}, our algorithm achieves a success rate of $96.04\%$ using only $3\%$ of the data as queried points. This performance is competitive with state-of-the-art active learning algorithms for the Salinas dataset (see \cite{cloninger2020cautious}).

\begin{algorithm}[htbp]
\caption{Signal Classification via Active learning (SCALe)}
\KwIn{Dataset $X \subset \mathbb{R}^{d}$, kernel degree $n$, threshold parameter $\Theta$, adjacency parameter $\eta$, step size $\eta_{\text{step}}$}
\KwOut{Predicted labels $\hat{y}$ for all points in $X$}

$\mathcal{A} \leftarrow \emptyset$ \;

    Apply PCA transformation $\mathbb{R}^d \to \mathbb{R}^{d_{min}}$ \;
    Project data onto unit hypersphere $\mathbb{S}^q$ \;
    Compute matrix $A_{ij} = \arccos(\langle x_i, x_j \rangle)$ \;
    Construct kernel matrix: $\Phi_n(\langle x_i, x_j \rangle)^2$ \;
    Prune values from matrix by $\mathcal{G}_n(\Theta)$ \;
\While{$\eta \leq \eta_{\max}$}{
    Build adjacency graph $G = (V, E)$ where $E = \{ (x_i, x_j) : A_{ij} < \eta \}$ \;
    Identify connected components $C_{\eta,\ell}\}_{\ell=1}^{K_n}$ ;

    \For{$\ell = 1$ \KwTo $K_n$}{
        \If{$C_{\eta,\ell} \cap \mathcal{A} = \emptyset$}{
            $x_i \leftarrow \underset{{x \in C_{\eta,\ell}}}{\argmax} \sum_{j=1}^{M} \Phi_n(\langle x, x_j \rangle)^2$ \;
            $\mathcal{A} \leftarrow \mathcal{A} \cup \{(x_i, f(x_i))\}$ \;
            {\small $\hat{y}(x_j) \leftarrow f(x_i)$ for all $x_j \in C_{\eta,\ell}$} \;
        }
        \ElseIf{$\forall x_j \in C_{\eta,\ell} \cap \mathcal{A}, f(x_j) = c_{\ell}$}{
            {\small $\hat{y}(x_j) \leftarrow c_{\ell}$ for all $x_j \in C_{\eta,\ell}$} \;
        }
    }

        $\eta \leftarrow \eta + \eta_{\text{step}}$ \;

}

Identify uncertain points $\mathcal{C}_{\text{uncertain}} \leftarrow X \setminus \bigcup_{\eta,\ell} C_{\eta,\ell}$ \;

$\mathcal{A}_k\gets \{x_j:\hat{y}(x_j)=k\}$ \;

\ForEach{$x_j \in \mathcal{C}_{\text{uncertain}}$}{
    $\hat{y}(x_j) \leftarrow \underset{k}{\argmax} \sum_{x_i \in \mathcal{A}_k} \Phi_{n,q}(\langle x_j, x_i \rangle)$
}

\Return $\hat{y}$ 
\end{algorithm}

\begin{figure}[h]
    \centering
    \includegraphics[width=0.4\textwidth]{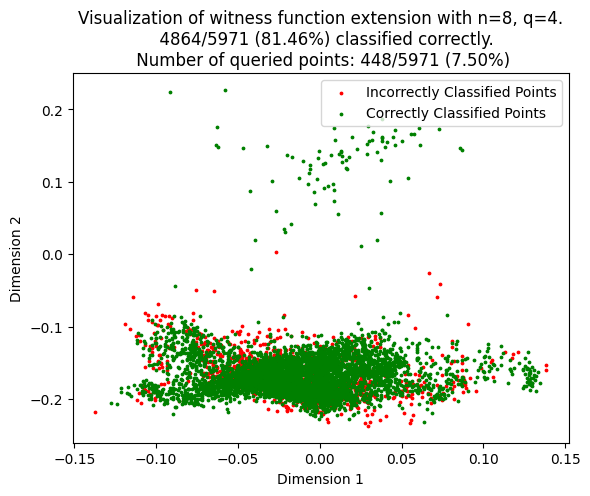}
    \caption{Indian Pines dataset.}
    \label{Indian}
\end{figure}

For the more challenging Indian Pines subset (see \Cref{Indian}), our algorithm achieves a success rate of $81.46\%$ using only $7.5\%$ of the data as queried points. This performance is also competitive with state-of-the-art active learning algorithms for the Salinas dataset (see \cite{cloninger2020cautious}).

\section{Conclusion}
In this paper, we introduced an active learning algorithm inspired by signal separation principles, demonstrating competitive performance on hyperspectral datasets with minimal labeled data. Our approach effectively identifies class supports even in the presence of overlapping distributions.Future work will focus on evaluating our algorithm’s generalizability across datasets from domains like medical imaging, remote sensing, and social networks to assess its adaptability to different classification tasks.
\newpage
\bibliographystyle{IEEEtran.bst}
\bibliography{Sampta2025bibliography}

\end{document}